\pdfoutput=1

\documentclass[11pt]{article}

\usepackage{acl}

\usepackage{times}
\usepackage{latexsym}
\usepackage[raggedrightboxes]{ragged2e}
\usepackage{nicematrix}
\usepackage[T1]{fontenc}

\usepackage[utf8]{inputenc}

\usepackage{microtype}

\usepackage{inconsolata}
\usepackage{graphicx}
\usepackage{booktabs}
\usepackage{tabularx}
\usepackage{multirow}
\usepackage{subcaption}
\usepackage{listings}
\usepackage{xcolor,colortbl}
\usepackage{transparent}  
\usepackage{nicematrix,tikz}
\usepackage[raggedrightboxes]{ragged2e}
\usepackage{enumitem}
\title{Discourse-Driven Evaluation: Unveiling Factual Inconsistency in Long Document Summarization}

\author{Yang Zhong \\
  Department of Computer Science \\
 University of Pittsburgh \\
  \texttt{yaz118@pitt.edu} \\\And
  Diane Litman \\
  Department of Computer Science and LRDC \\
 University of Pittsburgh \\
  \texttt{dlitman@pitt.edu} \\}

\begin{document}
\maketitle
\begin{abstract}
Detecting factual inconsistency for long document summarization remains challenging, given the complex structure of the source article and long summary length. In this work, we study factual inconsistency errors and connect them with a line of discourse analysis. We find that errors are more common in complex sentences and are associated with several discourse features. We propose a framework that decomposes long texts into discourse-inspired chunks and utilizes discourse information to better aggregate sentence-level scores predicted by natural language inference models. Our approach shows improved performance on top of different model baselines over several evaluation benchmarks, covering rich domains of texts, focusing on long document summarization. This underscores the significance of
incorporating discourse features
in developing models for scoring summaries for long document factual inconsistency.
\end{abstract}

\section{Introduction}

Current state-of-the-art summarization systems can generate fluent summaries; however, their ability to produce factually consistent summaries that adhere to the source content or world knowledge remains questionable. This phenomenon is known as \textbf{factual inconsistency}, one type of ``hallucination'' problem \cite{maynez-etal-2020-faithfulness, zhang2023languagemodelhallucinationssnowball, cao-wang-2021-cliff,kryscinski-etal-2020-evaluating, goyal-durrett-2021-annotating, cao-etal-2022-hallucinated}.  A rigorous line of research approaches this problem by developing models to detect unfaithful summary content, including utilizing pre-trained models such as natural language inference (NLI)  \cite{ kryscinski-etal-2020-evaluating, laban2022summac, zha-etal-2023-alignscore} and question answering (QA) \cite{scialom-etal-2021-questeval, fabbri-etal-2022-qafacteval} models. Such approaches are tested on rich benchmark datasets, such as \textsc{True} \cite{honovich-etal-2022-true}, \textsc{Summac} \cite{laban2022summac}, and \textsc{AggreFact} \cite{tang-etal-2023-understanding}, etc. 

However, such benchmark datasets only include short documents (< 1000 words) and summaries with a few sentences. While the methods mentioned above perform well with short texts, they struggle with longer documents \cite{schuster-etal-2022-stretching}. Recent NLI work addresses this by selecting the input and breaking down the summary. Lengthy summaries are split into individual sentences or more minor atomic claims, while small chunks of the source document are extracted as premises. This approach reduces the task to multiple short evaluations, which are then aggregated to provide a summary-level label \cite{zha-etal-2023-alignscore, zhang-etal-2024-fine, scire-etal-2024-fenice, yang2024fizz}. 

Out of the existing NLI-based methods, \textsc{AlignScore} demonstrated superior performance on multiple benchmarks. It breaks the input document into continuous chunks of text to tackle the input restriction. However, this exhaustive approach may break the structure of the context (section and paragraph split), thus reducing the chances that the summary sentence can be correctly verified with its factual consistency.  On the other hand, most factuality evaluation metrics aggregate the sentence-level aligning scores through averaging or selecting the minimum, disregarding that sentences are not equally important \cite{krishna-etal-2023-longeval}. For instance, people can remember the big picture more easily but struggle to retain low-level details when retelling a story. The natural questions would be: do system-generated summaries carry a similar pattern? If so, how can we utilize the text organization information to help detect the inconsistencies between the summary and the source document? 

In this work, we study the factual inconsistency problem through the lens of discourse analysis. By analyzing the structure (here we use Rhetorical Structure Theory (RST) \cite{MANNTHOMPSON+1988}) of the original articles and the summaries, we uncover the importance of preserving the article structure and studying the connections between discourse structure and the factual consistency of model-generated summaries. Our analysis shows that complex sentences built by multiple elementary discourse units (EDUs, the basic units used in the discourse theory) have a higher chance of containing errors, and we also find several discourse features connected to the factual consistency of summary sentences.

Motivated by the analyses mentioned above, we propose a new evaluation method, \textsc{StructScore}, based on the NLI-based approaches to better detect factual inconsistency. Our algorithm includes two steps:  (1) leveraging the discourse information when aggregating the sentence-level alignment scores of the target summary and (2) decomposing the long input article into multiple discourse-inspired chunks. We tested our proposed approach on multiple document summarization benchmarks, including \textsc{AggreFact-FtSOTA} split, \textsc{DiverSumm}, \textsc{LongSciVerify}, \textsc{LongEval}, and a non-scientific domain dataset \textsc{LegalSumm} with a focus on long document summarization. Our proposed approach obtained a performance gain on multiple tasks.\footnote{Our models and model outputs are publicly available at \url{https://github.com/cs329yangzhong/discourse-driven-summary-factuality-evaluation}} 

To sum up, two research questions are addressed:
1. How and what discourse features are connected to the factual inconsistency evaluation?
2. Can our discourse-inspired approach improve the detection performance on long document summarization?

\section{Related Work}

\paragraph{Factual Inconsistency Detection in Long Document Summarization}

Research on automatic factual inconsistency evaluation metrics and resources for long document summarization is limited.
Recently, \citet{Koh2022AnES} surveyed the progress of long document summarization evaluation and called for better metrics and corpora to evaluate long document summaries. \citet{koh-etal-2022-far} released annotated model-generated summaries assessing factual consistency at the
\textbf{sentence} and \textbf{summary} levels for GovReport \cite{huang-etal-2021-efficient} and arXiv \cite{cohan-etal-2018-discourse}. Furthermore, \citet{bishop-etal-2024-longdocfactscore-evaluating} and \citet{zhang-etal-2024-fine} introduced benchmarks of \textsc{LongSciVerify} and \textsc{DiverSumm} that cover diverse domains respectively, and further proposed different frameworks to utilize the context of source sentences for evaluating the factual consistency of generated summaries. However, their approaches relied on extracting context through computing similarities with the summary sentence. The summary-level score is a simple average of all sentence-level predictions. \textit{Our work analyzed a subset of \textsc{DiverSumm} and \textsc{AggreFact} \cite{tang-etal-2023-understanding} that have sentence-level factual inconsistency types and introduced a generalizable approach to better detect such inconsistency errors across domains.}


\paragraph{Aggregation of Sentence-level Evaluations}
Text summaries are usually composed of multiple sentences. Most factual inconsistency evaluation metrics first compute the sentence-level scores for individual summaries, then aggregate them by either \textbf{soft aggregation} in computing the 
\textbf{unweighted-average} \cite{ zha-etal-2023-alignscore, glover-etal-2022-revisiting,scire-etal-2024-fenice,zhang-etal-2024-fine} or  \textbf{hard aggregation} with the minimum score \cite{ schuster-etal-2022-stretching,yang2024fizz}. ver, these approaches were primarily validated on older benchmarks, consisting of shorter texts (a few hundred input words and summaries of 2-3 sentences). There lacks a systematic study in the context of long document summarization. \textit{Our work dives into the discourse structure of system-generated summaries with span/sentence-level factuality annotations. We introduce a discourse-inspired re-weighting algorithm to calibrate the scores.}
\vspace{-1mm}

\paragraph{Discourse-assisted Text Summarization} Discourse factors have been known to play an important role in the summarization task \cite{ono-etal-1994-abstract, Marcu1998tobuild, kikuchi-etal-2014-single, xu-etal-2020-discourse, hewett-stede-2022-extractive, pu-etal-2023-incorporating}. \citet{louis-etal-2010-discourse} conducted comprehensive experiments to examine the power of different discourse features for context selection. We carry a similar analysis but focus on summary sentences that contain factual inconsistency errors. On adjusting the weight of EDUs, \citet{huber-etal-2021-w} proposed a weighted RST style discourse framework that derives the discourse units' continuous weights from auxiliary summarization task \cite{xiao-etal-2021-predicting}. Differently, our re-weighting algorithm is built on top of the trained parser's parsed discourse tree and applies to the final aggregation of scores. \textit{To the best of our knowledge, our work is the first that studies the connections between RST discourse structure and the factual consistency of model-generated summaries.}

\begin{table*}[th!]
\scriptsize
    \centering
       \setlength\columnsep{1pt}
    \begin{NiceTabular}{l|crrrrr}
    \toprule
      \textbf{Dataset}  & \textbf{Sum.Task} &  \textbf{Size}  & \textbf{Doc.Word} & \textbf{Doc.Sent} & \textbf{Sum.Sent} & \textbf{Sum.Word}  \\
      \midrule
       \multirow{2}{*}{\textsc{AggreFact-FtSOTA}} &XSum \cite{tang-etal-2023-understanding} & 558 & 360.54 & 16.09 & 1.01 & 20.09\\
         & CNNDM \cite{tang-etal-2023-understanding} & 559 & 518.85 & 23.31& 2.72 & 52.21 \\
    \midrule
         \multirow{5}{*}{\textsc{DiverSumm}} & Multi-news  \cite{fabbri-etal-2019-multi} &  90& 669.20 & 27.2 & 6.81 & 152.20 \\
         & QMSUM \cite{zhong-etal-2021-qmsum}  & 90 & 1138.72 &72.80 & 3.04 & 65.22\\
         & Government \cite{huang-etal-2021-efficient} & 147 & 2008.16 & 71.35 & 15.1 & 391.22\\
           & ArXiv \cite{cohan-etal-2018-discourse} & 146& 4406.99 & 195.18 & 6.18 & 149.70 \\
           & ChemSumm \cite{adams-etal-2023-desired}  & 90 & 4612.40 & 188.80 & 7.36& 172.79\\

    \midrule
    \multirow{2}{*}{\textsc{LongSciVerify}} & PubMed \cite{cohan-etal-2018-discourse} & 45 & 3776.80 & 125.00 & 8.60& 225.60 \\
    & ArXiv \cite{cohan-etal-2018-discourse} & 45 & 6236.40 & 282.93 & 7.28 & 210.93\\

    \midrule
    \textsc{LongEval} & PubMed \cite{krishna-etal-2023-longeval} & 40 & 3158.35 & 110.00 & 10.38 & 193.55 \\
    \midrule
    \textsc{LegalSumm} & Legal Opinions \cite{elaraby-etal-2023-towards} & 50 & 2873.87 & 115.64 & 8.36 & 208.28 \\
    \bottomrule
    \end{NiceTabular}
    \caption{Summary-level task statistics on \textsc{AggreFact-FtSOTA}, \textsc{DiverSumm}, \textsc{LongSciVerify}, \textsc{LongEval} and \textsc{LegalSUMM}. We report the number of annotated doc-summary pairs of the test split (Size), document length in the average
number of words (Doc.Word) and the average number of sentences (Doc.Sent), summary length in the average number of sentences (Sum.Sent), and words (Sum.Word). }\label{tab:detection_dataset_analyses}

\end{table*}

\section{Datasets}
This section describes the datasets used to explore our research questions. We begin with the discourse analysis dataset, which includes sentence-level fine-grained labels of errors introduced in \citet{pagnoni-etal-2021-understanding}, enabling systematic analysis of the relationships between different features and their labels. We then discuss the benchmark datasets, which provide summary-level labels in either binary or continuous scores, and evaluate our approach and baselines on them.

\paragraph{Discourse Analysis Dataset}\label{token_level_dataset}
Our discourse analysis harnessed the subsets of \textsc{ArXiv} and \textsc{GovReport} from \textsc{DiverSumm} \cite{zhang-etal-2024-fine}, which come with annotated sentence-level errors labels. Following \citet{zhang-etal-2024-fine}, we denote it as \textsc{DiverSumm-sent}.  It covers 293 document-summary pairs of which 3138 summary sentences have sentence-level annotations.\footnote{We include analysis of the short document summarization datasets in Appendix \ref{appendix:short_analysis}.}

\paragraph{Summary-level Factuality Detection Datasets}
We test on the \textsc{AggreFact-FtSOTA} split \cite{tang-etal-2023-understanding},
\textsc{DiverSUMM} \cite{zhang-etal-2024-fine}, \textsc{LongSciVerify} and \textsc{LongEval} from \citet{bishop-etal-2024-longdocfactscore-evaluating}. We additionally collect \textsc{LegalSumm}, a legal summarization dataset, which covers model-generated summaries from the CanLII (Canadian Legal Information
Institute) dataset \cite{xu2021, elaraby-etal-2023-towards} with document-level factuality labels annotated by legal experts.\footnote{We provide details of the dataset in Appendix \ref{appendix:legalsumm_detail}.} Table \ref{tab:detection_dataset_analyses} presents a careful comparison of datasets from different perspectives. We conduct analysis on the document's structure in \S\ref{sec:document_structure} using these datasets. Except for \textsc{AggreFact}, all remaining datasets are focused on long documents and summary pairs.

\vspace{-2mm}
\section{Discourse Analysis}\label{sec:discourse_analysis}

\begin{figure*}[t!]
    \centering
    \includegraphics[width=0.99\textwidth]{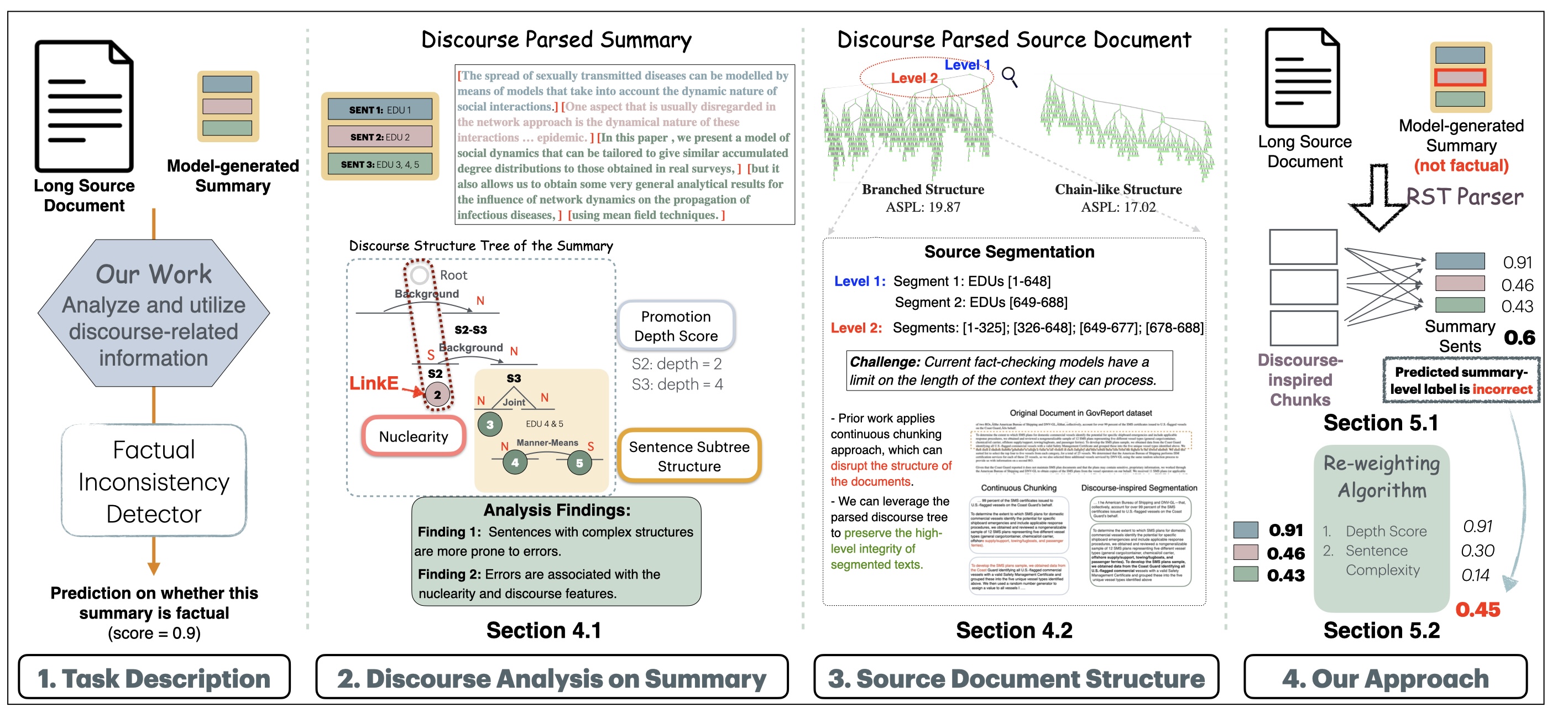}
    \caption{Our proposed approach to faithfulness inconsistency detection utilizes findings from discourse analysis. We first conduct discourse analysis on parsed summary sentences (\S \ref{sec:summary_error_analysis}) and exploit the source document's discourse structure (\S \ref{sec:document_structure}). Motivated by the findings, our proposed approach is introduced in \S  \ref{sec:source_segment} and \S \ref{sec:reweight_algorithm}.  }
    \label{fig:discourse_inspired_detection}
\end{figure*}
\paragraph{Preliminaries}

Discourse analysis with Rhetorical Structure Theory (RST) is helpful for different downstream tasks, such as argument mining \cite{peldszus_stede_2016a, hewett-etal-2019-utility}, text simplification \cite{Zhong_Jiang_Xu_Li_2020},  AI-generated text detection \cite{kim2024threads}, and summarization \cite{Marcu1998tobuild, xu-etal-2020-discourse}. \textbf{RST} predicts tree structures on the grounds of underlying coherence relations
that are primarily defined in speaker intentions \cite{MANNTHOMPSON+1988}. 
The discourse tree comprises lower-level Elementary Discourse Units (EDUs), each corresponding to a phrase within a sentence. These units are then integrated into more complex structures, such as sentences and paragraphs, to form the full discourse tree. Discourse labels (i.e., elaboration, contrast, condition, etc.) are assigned as the relation between nodes. Additionally, a nuclearity attribute is assigned to every node of the discourse tree, aiming to encode the relative importance between the pairs of sub-trees (nucleus roughly implying primary importance and a satellite means supplemental).


We first parse the summaries from the datasets as mentioned earlier in Section \ref{token_level_dataset} with an open-sourced DMRST model \cite{liu-etal-2021-dmrst}, following similar work which utilizes the same model for discourse parsing \cite{adams-etal-2023-generating,pu-etal-2023-incorporating, kim2024threads}. In the following paragraphs, we propose and verify multiple hypotheses that inspired our discourse-structure-aware factual inconsistency detection approach. Figure \ref{fig:discourse_inspired_detection} summarizes our findings in \S \ref{sec:summary_error_analysis} and \S \ref{sec:document_structure}.

\subsection{Discourse Analysis on Summary Errors}\label{sec:summary_error_analysis}

\begin{table}[]
\renewcommand{\arraystretch}{0.6}
\scriptsize

    \centering
    \setlength\columnsep{3pt}
    \begin{tabular}{c|c|c|c}
    
    \toprule
    \textbf{Error} & \multicolumn{3}{c}{\textbf{Discourse Subtree Depth}}\\
    \cmidrule{2-4}
    & -1 & 0  &  >= 1 \\
 &  (split link) &  (1 edu) & shallow/deep trees \\
 \midrule 
 GramE &  6\% & 28\% & 66\% \\
 \midrule  
  LinkE & 14\% & 23\% & 63\%\\
 \midrule  
  OutE & 15\% & 13\% & 72\%\\
  \midrule
   EntE & 11\% & 10\% & 79\%\\
 \midrule  
  PreE & 20\% & 13\% & 67\%\\
 \midrule  
  CorefE & 11\% & 0\% & 89\%\\
  \midrule 
 CircE &  8\% & 8\% & 84\% \\
 \midrule
 NoE & 8\% & 23\% & 69\% \\
 
 \bottomrule
    \end{tabular}
    \caption{The distribution depths of discourse subtrees of a sentence that are not factually consistent (depth of sub-tree) in \textsc{DiverSumm-Sent}. ``-1'' means the original sentence belongs to two sub-trees. Appendix \ref{sec:appendix_discourse_test} includes details of error types. }
    \label{tab: subtree_struct}
\end{table}

\paragraph{Finding 1:  Errors are located in sentences with dense  discourse tree (more EDUs)}
RST can capture the salience of a
sentence with respect to its role in the larger context. Prior work finds that the salience of a unit or sentence does not strictly follow the linear order of appearance in the document but is more indicative through its depth in the tree \cite{Zhong_Jiang_Xu_Li_2020}. We consider the depth of the current sentence in
the RST tree of the document (viewing each sentence as a
discourse unit). We also noted that, at times, the original summaries' sentences are broken into parts and span two discourse subtrees (i.e., a sentence covers EDUs 24-28, while the parsing tree's subtrees are ``22-25''', ``26-28''). In this case, we approximate the depth of the sentence by computing the square root of the absolute distance of min and max EDUs, i.e., in the above case, the depth is computed as $\sqrt{(28-24)}=2$.\footnote{We assume that the discourse tree is nearly binary, with each node having two children.}

We additionally studied the distribution of the tree structure of sentences with errors. The hypothesis is that several errors will likely appear in sentences with complex structures (more EDU units and dense trees). As shown in Table \ref{tab: subtree_struct}, sentences containing factual inconsistency errors are generally more complicated and cover multiple discourse units. It is worth noting that the case of ``-1'' means the sentence is deeply intervened with its neighboring sentences, and the discourse parser fails to segment it independently. One example is illustrated in the summary of Figure \ref{fig:discourse_inspired_detection}, where Sentence 3 (S3) contains three EDU segments, making it more complex than the other two sentences.

\paragraph{Finding 2: Errors are associated with the nuclearity and related discourse features}
We further analyze the distribution of nuclearity and different discourse features of sentences containing errors from the \textsc{DiverSumm-Sent} dataset. We observe that a greater number serve as satellites within the discourse relation (62\%) for sentences comprising a single Elementary Discourse Unit (EDU). 

\begin{table}[]
    \centering
    \small
    \begin{NiceTabular}{l|c|l}
    \toprule
    \textbf{RST features}     &  \textbf{t-stat} & \textbf{p-value}\\
    \midrule
      Ono penalty \cite{ono-etal-1994-abstract}    & 1.606 & 0.1089 \\
      Depth score \cite{Marcu1998tobuild} & \textbf{-9.084} &0.0000\\
    Promotion score \cite{Marcu1998tobuild} & -0.828 &0.4083\\
    \midrule
    Normalized Ono penalty  & \textbf{2.160} &0.0314\\
      Normalized depth score& \textbf{-8.919} &0.0000\\
        Normalized promotion score & -0.303 & 0.7617\\
        
    \bottomrule
    \end{NiceTabular}
    \caption{Two-sided t-test of significant RST-based features comparing sentences with factual inconsistency errors to consistent ones in \textsc{DiverSumm-Sent}. We report the test statistics and significance levels. The original and normalized depth scores and the normalized penalty scores are significant (p-value <= 0.05). Fine-grained per error-type results are in Table \ref{tab:rst_result_all_errors} of Appendix \ref{sec:appendix_discourse_test}.}
    \label{tab:rst_feature_diversumm}
\end{table}

We calculated several discourse feature scores: the penalty score (Ono penalty) as defined in \citet{ono-etal-1994-abstract}, the maximum depth score (Depth score) \cite{Marcu1998tobuild}, and the promotion score \cite{Marcu1998tobuild}.\footnote{Details of feature scores are in Appendix \ref{appendix:disocurse_feature_details}.} The penalty score accounts for
the number of satellite nodes found on the path from
the tree's root to that EDU. The depth score is determined by the proximity of an EDU's highest promotion to the tree's root. The highest promotion refers to the closest node to the root, including the EDU within its promotion set. The promotion score quantifies the salience of an EDU based on how many levels it has been promoted through within the tree structure. We compute both unnormalized and normalized versions (with the max tree depth) for the above three scores. As shown in Table \ref{tab:rst_feature_diversumm}, we find significant differences in the distributions of depth score. We normalize the Ono penalty and depth score between factually consistent and inconsistent sentences and will include them in our proposed approach.


\subsection{Document Structure}\label{sec:document_structure}
\begin{figure}
    \centering
    \includegraphics[width=0.97\linewidth]{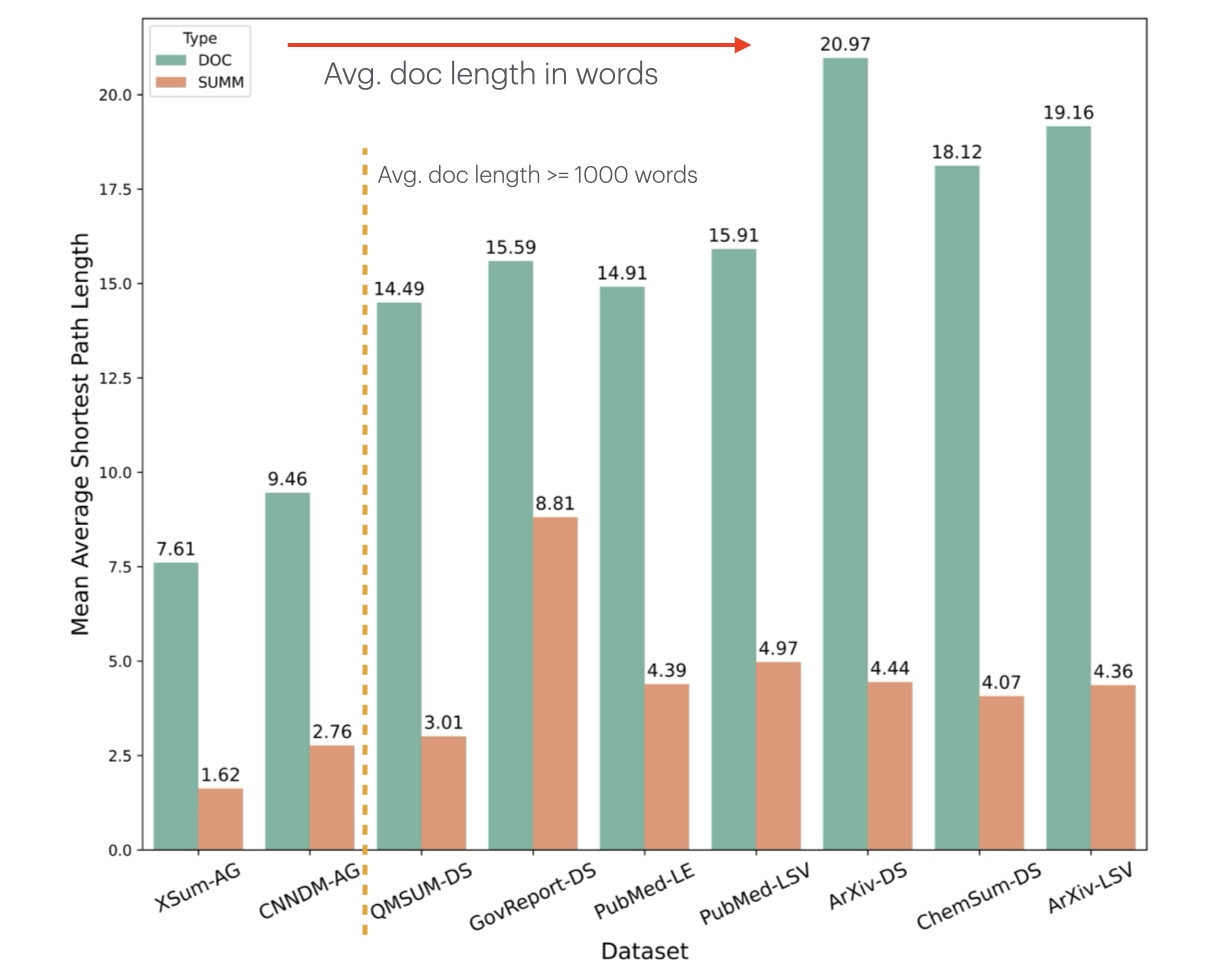}
    \caption{Average shortest path length per dataset for document and summary discourse trees. We sort the dataset by the average length of the document, finding that longer document-summary (DOC, SUMM) pairs would be more branched, and their summaries are also complicated. AG, DS, LSV, and LE refer to \textsc{AggreFact FtSOTA}, \textsc{DiverSumm}, \textsc{LongSciVerify} and \textsc{LongEval} respectively.}
    \label{fig:aspl_fig}
\end{figure}

 We further analyze the structure of parsed discourse trees for both documents and summaries of different datasets. We assume that the linguistic structure of discourse can change depending on factors such as the writing style, domain, and depth of reasoning of texts. To check whether the
structures are evenly branched or follow a more sequential pattern, we measure a document graph's average shortest path length \cite{kim2024threads}. The intuition is that linear or chain-like graphs tend to have shorter average shortest path lengths (ASPL), reflecting the linear pattern. Meanwhile, branched structures would have a longer ASPL, given the spread nature of nodes. As shown in Fig \ref{fig:aspl_fig}, for long document datasets (the last seven datasets), the source documents' ASPL is longer than the news articles such as CNN/DM and XSUM.\footnote{We exclude Multi-news and \textsc{LegalSumm}, as the former dataset's source text is composed of multiple news articles and the latter comes with split section structures, making the ASPL reporting less accurate.} In the meantime, longer summaries also carry evenly branched complex structures compared to short news summaries. While mainstream research segments long source texts into continuous chunks with limited window size, we argue that this disrupts the original structure of texts, leading to information loss.\footnote{See Appendix \ref{appendix:segmentation_examples} for examples.} We propose utilizing the tree structure and constructing the segments based on level traversals to preserve the high-level segmentation. 

\section{StructScore}
In this section, we describe the \textsc{StructScore} framework. The lower right part of Figure \ref{fig:discourse_inspired_detection} presents motivations for each module.
\subsection{Tree-structure Inspired Weighting Algorithm}\label{sec:reweight_algorithm}
Prior work \cite{zha-etal-2023-alignscore, scire-etal-2024-fenice} computes the aggregated summary-level prediction on factual consistency score by picking the minimum sentence-level score or selecting the average. However, as indicated in Section \ref{sec:summary_error_analysis}, EDUs with different discourse relations and structures can be weighted differently. We thus propose to re-weigh the sentences based on the features of the discourse. 

First, we examine the sentence's nuclearity and the associated discourse features within the discourse tree. As found in Table \ref{tab:rst_feature_diversumm}, the normalized depth score, which utilizes the given node's nuclearity and the tree structure, is significantly different given the existence of factual inconsistency errors (p-value < 0.00001), where inconsistent sentences have a lower normalized depth score (Finding 2 in \S \ref{sec:summary_error_analysis}).\footnote{Among the three significant features, we use the normalized depth score to ensure consistent scaling. Our preliminary results also indicate that the normalized Ono penalty score did not enhance the dev set performance as much. } Based on this finding, we decided to increase the weight of the alignment score for sentences with lower depth scores within their parsed tree. Since NLI methods generate scores within a 0-1 range, we apply an exponent to appropriately scale these scores.
Let \( x_i \) be the computed normalized depth score of a summary sentence, \(s_i\) the original computed aligning score,  and \( \overline{x}_{1:j} \) the mean of all depth scores from \( x_1 \) to \( x_j \) in the summary with length j. The function to re-weight the aligning score \( f(s_i) \) can be defined as follows:
\[
f(s_i) = 
s_i^{1+ (\overline{x}_{1:j} - x_i)} 
\]
Secondly, observing that sentences that contain connective EDUs or have complicated discourse structures with more EDUs are more likely to contain errors (Finding 1 in \S \ref{sec:summary_error_analysis}), we propose scaling the score by selecting an appropriate exponent, given that the original score falls within the range of 0 to 1. We apply a tuning factor $\alpha$ on the discourse sub-tree height for the summary sentence $sent_i$: 
\[s_i^{*} = f(s_i)^{1+({height-subtree}(sent_i)*\alpha)}\] We conduct ablation studies on these two components in \S \ref{sec:abalation_diversumm}. We search for the best parameters on a held-out dev set of \textsc{DiverSumm} and keep the same across other datasets.

\subsection{Source Document Segmentation}\label{sec:source_segment}

We parse the original article with the RST parser and break the long documents into linear segments. This approach differs from prior work, which either applies a fixed window or selects a few context sentences surrounding a given source sentence. Motivated by findings from \S \ref{sec:document_structure}, we follow the below approach:
(1) If the parser fails, we will use the document structure (paragraph/sentence hierarchies) to group by the neighboring sentences. We then follow the naive chunking approach in \textsc{AlignScore} (window size 350) to prepare the input. 
(2) If the parsing is successful, we will extract the segmentation from the discourse tree up to level N. For instance, in the top-right of Figure \ref{fig:discourse_inspired_detection}, an original article has EDU segments (1-688), and the root of the RST tree is split into 1-648 and 649-688; we will adopt this segmentation. We apply the chunking approach outlined previously for segments that exceed the \textsc{AlignScore} model's context capacity. On the second level, we break (1-648) into (1-325) and (326-648), while the remainder are also broken into smaller chunks. Since the RST parser could break long sentences into multiple EDUs, we have additional post-processing to map the EDUs back to the source sentences.

\section{Experimental Details}
\newcommand{\dashrule}[1][black]{%
  \color{#1}\rule[\dimexpr.5ex-.2pt]{1pt}{.4pt}\xleaders\hbox{\rule{1pt}{0pt}\rule[\dimexpr.5ex-.2pt]{3pt}{.4pt}}\hfill\kern0pt%
}

We adapt mainstream evaluation setups for each benchmark. For \textsc{DiverSumm}, we apply an 80/20 test/dev split by stratifying the labels for each subtask. For \textsc{AggreFact}, we use their released val/test split. For \textsc{LongSciVerify}, \textsc{LongEval} and \textsc{LegalSumm}, we use them as test sets.

\newcommand{\gray}[1]{\textcolor{gray}{#1}}
\newcolumntype{?}{!{\vrule width 1pt}}
\begin{table*}[t!]
\scriptsize
\centering
\setlength{\tabcolsep}{3pt}

\begin{tabular}{llll?llllc|l?ll?c|c}
 \toprule
\multirow{2}{*}{\shortstack{\textbf{ID} }} & \multirow{2}{*}{\shortstack{\textbf{Evaluation Model} }} & \multicolumn{2}{c}{\textsc{\textbf{AggreFact}}} & \multicolumn{6}{c}{\textsc{\textbf{DiverSumm}}} &  \multicolumn{2}{c}{\textsc{\textbf{LSV}}} & \textsc{\textbf{LongEval}} & \textsc{\textbf{LegalS}}\\
\cmidrule(lr){3-4}\cmidrule{5-10} \cmidrule{11-14} 

&& XSM\textsubscript{AG} & CND\textsubscript{AG} &MNW  & QMS  & GOV 
 & AXV    & CSM & \textit{Macro-} & PUB & AXV & PUB &  \\
 & evaluation metric & \multicolumn{2}{c}{\textit{AUC}}  & \multicolumn{5}{c}{\textit{AUC}} & \textit{AVG} & \multicolumn{2}{c}{\textit{Kendal's $\tau$}} & \textit{Kendal's $\tau$} & AUC\\
& avg src. len & 360.54 & 518.85 & 669.20 & 1138.72   & 2008.16 & 4406.99 & 4612.40 & -- & 3776.80 & 6236.40 & 3158.35 & 2873.87  \\

\midrule

\multicolumn{12}{l}{\textit{\textbf{Baselines}}}\\
\midrule
\textbf{1} & \textbf{\textsc{LongDocFactScore}} & 50.47 &   65.27& \cellcolor{green!10}{61.20} & 40.69 & 83.52 & 65.36 & 60.06 & 62.17 &\cellcolor{green!10}{61.0}  & \cellcolor{green!10}{61.0} & 29.0 & 60.19 \\

\textbf{2} & \textbf{\textsc{MiniCheck-FT5}} & 75.04 &  72.62 & 48.68 & 45.31 & 70.26 & 61.77 & 52.93 & 55.79& 26.5 & 38.1 & 17.4 & 61.33\\

\textbf{3} &\textbf{GPT4o} & 75.36 & 70.47 &51.11 &\cellcolor{green!10}{70.22} & {86.81} & 67.78 & 61.53 & 67.49& 54.7 & 51.8 & \underline{51.2} & 67.71\\

\textbf{4} &\textbf{BeSpoke-MC-7B} & \cellcolor{green!10}{83.56} & 71.38  & 55.38 & \underline{65.42} &  82.83 & 75.07 & 63.43 & \underline{68.42} & 55.1 & \underline{57.9} & \cellcolor{green!10}{58.1} & 55.81 \\ 
\midrule

\multicolumn{12}{l}{\textit{{Apply our approach with different \textbf{baselines}(\textit{$\uparrow$} means improved the performance compared to the baseline with significance.)}}}\\

\midrule
\textbf{5} & \textbf{\textsc{AlignScore}} & {75.66} & 69.50 & 46.74 & 56.48 & {87.02} & 77.46 & 61.03 & 65.75 & 54.9 & 53.9 & 36.9 & {73.52}\\

 6 &   \hspace{3mm}+ re-weighting  & 75.67 & 69.20 &45.33 & 53.95 & 87.29$\uparrow$ & 81.15$\uparrow$ & 60.55 & 65.65 & 53.0 &  54.3$\uparrow$ & 34.8 & \cellcolor{green!10}{76.57}$\uparrow$\\
 \cmidrule{2-14}
 7 &\hspace{3mm}\textsc{+ Lv1 segment} & 76.23$\uparrow$ & 69.25\textsuperscript{$\dagger$} & 45.86\textsuperscript{$\dagger$}  & {61.25}$\uparrow$   & {86.74}\textsuperscript{$\dagger$} & 79.47$\uparrow$ & \underline{64.15}$\uparrow$ & 67.49$\uparrow$  & 51.9 & 52.8 & 43.6$\uparrow$ & 59.43\\
8& \hspace{3mm}\textsc{StructS-Lv1} & 76.20$\uparrow$ & 69.03 & 46.21\textsuperscript{$\dagger$}  & 60.06$\uparrow$ & 86.04  & {82.78}$\uparrow$   & \cellcolor{green!10}{64.47}$\uparrow$  & 67.91$\uparrow$  &  50.4 & 53.9\textsuperscript{$\dagger$} & 43.4$\uparrow$ & 59.81 \\
\cmidrule{2-14}

9& \hspace{3mm}\textsc{+ Lv2 segment}  & 74.27 & 70.30$\uparrow$  & 46.03\textsuperscript{$\dagger$}  & 55.74  & 85.10  & 76.79  & 63.11$\uparrow$  & 65.35 &  58.1$\uparrow$ & 51.1 & {43.9}$\uparrow$ & 67.05\\
10& \hspace{3mm}\textsc{StructS-Lv2}  &  74.28 & 69.85$\uparrow$&  45.33 & 51.86 & 85.65 & 80.00$\uparrow$ & 63.59$\uparrow$ & 65.29& 55.3$\uparrow$ & 54.1$\uparrow$ &  43.7$\uparrow$  & 64.00\\
 
\midrule
\midrule

\textbf{11}& \textbf{\textsc{MC-FT5 (sent)}} &  {79.62} & 70.95 & \underline{57.67} & 60.66&  83.24 & 78.66 & 59.74 & 67.99 & 55.7 & 52.7 & 30.2 & 61.14\\
12&\hspace{3mm}+ re-weighting & \underline{79.73} & 70.76\textsuperscript{$\dagger$} & 56.79 & 60.36\textsuperscript{$\dagger$}  & 84.75$\uparrow$ & 79.38$\uparrow$ & 60.06$\uparrow$ & {68.27}$\uparrow$  & 52.8 & 55.1$\uparrow$ & 31.4$\uparrow$  & 59.81\\
\cmidrule{2-14}
13&\hspace{3mm}\textsc{+ Lv1 segment} & 77.84 & \cellcolor{green!10}{73.48$\uparrow$} & 44.80 & 61.10$\uparrow$ & \underline{87.50}$\uparrow$ &\underline{85.22}$\uparrow$ & 63.59$\uparrow$ & \cellcolor{green!10}68.44$\uparrow$   & 57.5$\uparrow$ & 51.4 &  33.0$\uparrow$ & 68.95$\uparrow$ \\
14&\hspace{3mm}\textsc{StructS-Lv1} & 76.75 & \cellcolor{green!10}{73.40$\uparrow$}& 38.45 & 60.66\textsuperscript{$\dagger$}  & \cellcolor{green!10}88.05$\uparrow$ & \cellcolor{green!10}{86.32}$\uparrow$ & 63.11$\uparrow$ & 67.31 &  56.2$\uparrow$ & 53.8$\uparrow$ & 30.7$\uparrow$ & 72.57$\uparrow$\\
\cmidrule{2-14}
15&\hspace{3mm}\textsc{+ Lv2 segment} & 73.70&  72.30$\uparrow$ & 47.80 & 57.53 & 86.26$\uparrow$& 83.73$\uparrow$ & 62.07$\uparrow$ & 67.48 & 56.0$\uparrow$ & 52.9$\uparrow$&  35.6$\uparrow$ & 72.57$\uparrow$\\
16&\hspace{3mm}\textsc{StructS-Lv2} & 71.31 & 72.30$\uparrow$ & 41.27 & 59.02 & 87.16$\uparrow$ & 84.78$\uparrow$ & 61.75$\uparrow$ & 66.80 & 53.4 & 54.2$\uparrow$ & 33.0$\uparrow$ & \underline{73.71}$\uparrow$\\
\midrule 
\midrule
\textbf{17}&\textbf{\textsc{InFusE}} &68.48 & {72.52} & 54.14 & 39.64 & 84.41 & 68.13 & 57.82 & 60.83 &  \underline{59.4} & 55.9 & 36.9 & 63.43\\
18&+ re-weighting& 67.30 & {72.37} & 53.44 & 40.54$\uparrow$ & 84.68$\uparrow$ & 74.31$\uparrow$ & 59.82$\uparrow$ & 62.56$\uparrow$ &  58.3 & {56.3}$\uparrow$ &  34.6 & 66.29$\uparrow$\\

\bottomrule

\end{tabular}
\caption{Results for all summarization tasks in \textsc{AggreFact-FtSOTA} (\textsc{AggreFact}), \textsc{DiverSumm}, \textsc{LongSciVerify} (LSV), \textsc{LongEval} and \textsc{LegalSumm} (LegalS).  In \textsc{DiverSumm},  CSM, MNW, QMS, AXV, and GOV refer to ChemSum, MultiNews, QMSUM, ArXiv, and
GovReport. We also report the macro-average of \textsc{DiverSumm} AUC. We highlight the \colorbox{green!10}{best} performed approach where multiple greens indicate systems indistinguishable from the best
according to a paired bootstrap test with p-value < 0.05, and the \underline{second-best} system for each column. The seven baseline models are \textbf{bolded}. Cells with \textsuperscript{$\dagger$} mean the result is {indistinguishable} from the raw baseline according to the bootstrap test. We report the average of 3 runs for GPT4o, given the randomness in LLM inference. 
}\label{tab:aggrefact_diversum_res}
\end{table*}

\paragraph{Baselines}
One of our baselines is \textbf{\textsc{AlignScore}} \cite{zha-etal-2023-alignscore}, an NLI-based metric that computes the aggregated inference score between a source article and generated summaries. We included \textbf{\textsc{Infuse}} \cite{zhang-etal-2024-fine}, which sets the SOTA on \textsc{DiverSumm},  \textbf{\textsc{MiniCheck FT5}} (MiniCheck-FlanT5 checkpoints) \cite{tang2024minicheck} that is a best-performing non-LLM fact-checker over multiple benchmarks, and \textbf{\textsc{LongDocFactScore}} \cite{bishop-etal-2024-longdocfactscore-evaluating} which claimed to work well on factuality validation of lengthy scientific article summaries. Our experiment notes that \textsc{MiniCheck} did not work well over long summaries due to its design objectives of short-statement fact-checking. We thus introduce \textbf{\textsc{MC-FT5 (SENT)}}, which computes the individual summary sentences' scores using \textsc{MiniCheck} and reports their average as the final summary score. We additionally include the \textbf{GPT4o} \cite{openai2024gpt4technicalreport} as the LLM fact-checker, using a prompt adopted from \citet{tang2024minicheck} (see Table \ref{tab:gpt4o_prompt} in Appendix \ref{appendix:implementation_details}). Lastly, we include \textbf{Llama-3.1-BeSpoke-MiniCheck-7B (BeSpoke-MC-7B)}\footnote{\url{https://huggingface.co/bespokelabs/Bespoke-MiniCheck-7B}}, the SOTA fact-checking model on the LLM-AggreFact benchmark \cite{tang2024minicheck}. Unless otherwise noted, we reran the baseline models on our datasets using the original authors' released code and checkpoints. Implementation details are provided in Appendix \ref{appendix:implementation_details}.
\paragraph{Our Approach}

We re-utilized  baseline models to compute the scores between context chunks and summary sentences, including \textsc{AlignScore} \cite{zha-etal-2023-alignscore}, \textsc{MiniCheck-FT5} (SENT) and \textsc{InfUsE} \cite{zhang-etal-2024-fine}, and experimented with below settings to apply our proposed approaches:
\setlist{nolistsep}
\begin{itemize}
\itemsep0em 
    \item + re-weighting: we apply the discourse-inspired re-weighting algorithm to adjust the sentence-level scores. We tune the factor $\alpha$ on height-subtree weighting as 1 over the validation set of \textsc{DiverSumm} and apply it to other benchmark datasets.
    \item + LvN \textsc{Segment}: Instead of using the default chunking approach, we segmented the source documents with the algorithms introduced in Sec. \ref{sec:source_segment} with different levels of granularity. 
    \item \textsc{StructS}-LvN: Combining top two methods.
\end{itemize}

The reweighting and segmentation can not be applied to \textsc{LongDocFactScore}, as it produced negative scores on all enumeration of source-target sentence pairs, which does not utilize the structural information. \textsc{InfUsE} utilizes the ranked list of entailment scores for all document sentences associated with each summary sentence. Thus, the segmentation approach does not affect.  
\vspace{-2.5mm}
\paragraph{Evaluation Metrics}

For experiments with {\textsc{AggreFact-FtSOTA}, \textsc{DiverSumm}} and \textsc{LegalSumm}, following \citet{laban2022summac, zhang-etal-2024-fine}, we adopt ROCAUC which measures classification
performance with varied thresholds as our evaluation metric.
On \textsc{LongSciVerify} and \textsc{LongEval}, we report Kendall's Tau $\tau$, following \citet{bishop-etal-2024-longdocfactscore-evaluating}.

\section{Results}\label{sec:result}
\textbf{Overall Performance} Table \ref{tab:aggrefact_diversum_res} presents our main results with detailed setups. Overall,  our proposed approach (with different combinations of re-weighting and segmentation settings) achieves the best or second best across \textsc{AggreFact}, most of \textsc{DiverSumm} and \textsc{LegalSumm} (\textsc{LegalS}).
Compared to top-performed LLM-based models (rows 3,4), our approach outperforms in 7 out of 11 datasets, with significant improvements on GOV, AXV, CSM, and \textsc{LegalSumm}.\footnote{More discussions on strong baselines in Appendix \ref{appendix:more_discussion}.}
The rest of the section addresses the following research questions: \textbf{RQ1:} Can the re-weighting algorithm help improve the models' performance?  \textbf{RQ2}: How does source document segmentation impact factual inconsistency detection? \textbf{RQ3}: How does combining both in \textsc{StructScore} perform?
\vspace{-1mm}
\paragraph{RQ1.}\label{sec:abalation_diversumm} \textit{We observe that the re-weighting algorithm improves prediction performance on different baselines (rows 5-6, 11-12, 17-18).} For long source documents, the re-weighting approach consistently improves or closely matches GOV, AXV, CSM splits in \textsc{DiverSumm} and the AXV split in \textsc{LongSciVerify} (LSV-AXV) and \textsc{LegalS} performance. Noticeably, \textsc{AlignScore} with reweighting scored the best on LegalS. On the other hand,  for both XSM and CND in \textsc{AggreFact-FtSOTA}, the re-weighting algorithm does not help much. We posit that the short summary length (1-3 sentences) has minimally structured information, so the scores will not change much. For MNW and QMS, the short summaries in QMS (averaging 3 sentences) reduce the effectiveness of the re-weighting algorithm. Moreover, MNW's non-factual sentences often receive high prediction scores, which our re-weighting approach tends to amplify, leading to a drop in performance. We also observe a slight performance drop on LSV-PUB and \textsc{LongEval-PUB} for \textsc{AlignScore} and \textsc{InfUsE}, potentially due to the different document structure of scientific articles from the medical domain. These observations also suggest potential future work for a dynamic weighting algorithm based on the document structure and domain knowledge.
In Table \ref{tab:diversumm_ablation}, we ablate the two discourse factors from the re-weighting algorithm with our best baseline MC-FT5 (SENT) on a subset of long datasets, noticing both features are helpful, and the improvement in adding subtree height is greater.\footnote{We include a more complete table in Appendix \ref{appendix:ablation_study}.}
\begin{table}[ht!]
\scriptsize
\centering
\begin{NiceTabular}{lcccc}
 \toprule
\multirow{1}{*}{\shortstack{\textbf{Model} }}
 &   \textbf{GOV}  & \textbf{AXV}  & \textbf{CSM} & \textbf{LSV-AXV} \\
\midrule

{MC-FT5 (SENT)} & 83.24 & 78.66 & 59.74 & 52.73 \\

{\hspace{3mm}\textit{+ subtree height}} & 84.55 & 79.09 & 60.55 & 55.08 \\
{\hspace{3mm}\textit{+ depth score}} & 83.65 & 78.90 & 59.90 & 53.80 \\
 re-weighting  &  84.75 & 79.38 & 60.06& 55.08 \\

\bottomrule
\end{NiceTabular}
\caption{Ablation results on a subset of datasets from \textsc{DiverSumm} and \textsc{LongSciVerify}, the top and bottom rows are rows 11 and 12 in Table \ref{tab:aggrefact_diversum_res}.}\label{tab:diversumm_ablation}
\end{table}

\vspace{-1mm}
\paragraph{RQ2.} \textit{Applying document and discourse-structure-inspired approaches enhances performance across different baselines on long document summarization tasks.} We start by applying the level-1 and level-2 segmentation to preserve the document structures while segmenting at higher levels. For example, MC-FT5 (SENT) with \textsc{Lv1 Segment} (row 13) obtains the highest macro-average AUC on \textsc{DiverSumm}, a trend also observed with \textsc{ALignScore}. Specifically, comparing row 11 and row 13, the Lv1 \textsc{Segment} improved the model's performance on 7 of 8 long datasets from QMS to \textsc{LegalS} (i.e. 78.66 -> 85.22  and 83.24 -> 87.50  on AXV and GOV). However, the effect of fine-grained segmentation can vary depending on the document's length and structure. For instance, \textsc{AlignScore} in row 9 with Lv2 segment obtained better performance than Lv1 on LSV-PUB but was worse on QMS.
\vspace{-1mm}
\paragraph{RQ3.} \textit{Combining both approaches is not universally beneficial across all scenarios}. When both individual approaches contribute positively, the combined \textsc{StructS} generally achieves better performance, as seen in row 8 on AXV, CSM, and row 14 on AXV. However, when one component causes a performance drop, combining both often leads to weaker overall performance than the stronger component alone. For instance, on GOV, row 8 performs worse than row 5, likely due to the segmentation in row 7, making the model less accurate. Similarly, row 14 performs slightly better than row 11 on LSV-PUB, but row 13's improvement does not translate into better performance gains when combined with row 12.  Differences in evaluation metrics (AUC vs. correlation) and dataset sizes may also have influenced these outcomes (i.e., row 14 does not improve much on \textsc{LongEval}-PUB while rows 12 and 13 have larger gains).

\colorlet{green30}{green!20}
\colorlet{green50}{green!30}
\colorlet{green70}{green!40}
\colorlet{red30}{red!20}
\colorlet{red50}{red!30}
\colorlet{red70}{red!40}

\section{Conclusion}
In this work, we approach the factual inconsistency detection of long document summarization through the lens of discourse analysis. We find that discourse factors, with regard to sentence structure, are related to the factual consistency of sentences. We further propose a framework that leverages the source document structure and introduces re-weighting the sentence-level predictions on top of different NLI-based models, achieving performance gains across multiple long-document summarization evaluation datasets, including scientific articles and legal documents.

\section*{Acknowledgment}
This work is supported by the National Science
Foundation under Grant No. 2040490, by Amazon, and by the
  University of Pittsburgh Center for Research Computing through the resources
 provided. Specifically, this work used the HTC cluster, which is supported by NIH award number S10OD028483. We want to
thank the members of the Pitt PETAL group, Pitt
NLP group, Pitt AI Fairness and Law group, and anonymous reviewers for their
valuable comments in improving this work. We also thank Kevin D. Ashley and Morgan A. Gray for their help with data annotation. 

\section*{Limitations}
Our work contributes to the understanding of factual inconsistency errors in machine-generated summaries from the lens of discourse analysis. 
Here, we discuss several limitations.

\paragraph{Benefits of Discourse-driven Information}
Our current approach leaves discourse-relation information (i.e., the relation types such as Explanation, Elaboration, etc.) \textit{unused} on the system level; it would be interesting to utilize it to detect and resolve inconsistency errors. We also acknowledge the choices of our current re-weighting algorithm (exponential) can be further studied with more motivation. We selected the current configuration that performed best on the validation splits of \textsc{DiverSumm}, aligning well with linguistic analysis principles. We plan to extend the modeling into a more complex version, such as applying a graph neural network to the tree structure and including discourse relations for future work.

While large models like GPT-4 and future architectures may improve long-context understanding, recent research shows that LLMs still face challenges with hallucination detection and effectively utilizing extended contexts \cite{liu-etal-2024-lost,zhu2024haluevalwildevaluatinghallucinationslanguage, luo2024halludiallargescalebenchmarkautomatic}. Our contribution, which links linguistic cues to hallucination detection, remains crucial, especially for summarization tasks. We acknowledge that future LLMs with expanded context windows may no longer require input pre-processing. However, we argue that discourse-based segmentation will \textit{still offer critical benefits} (explicitly or implicitly by injecting the discourse analysis into the LLM through prompting or further finetuning). It will enhance the precision of factuality detection and evaluation by leveraging linguistic structures. Additionally, discourse information can provide interpretability to the model, which allows us to trace its evaluations to identifiable linguistic relations and features, which are still lacking in LLMs. In fact, our experimental results with BeSpoke-MC-7B, the SOTA fact-checking model, support the assumption that LLM alone still struggles with the factuality evaluation of long summaries.

\paragraph{Computation Cost}
Our approach's only additional computation cost is running the discourse parser on the source document and the target summary. The DMRST parser \cite{liu-etal-2021-dmrst} can be run on both CPU and GPU, and the inference speed is fast (the full test set of \textsc{DiverSumm} can be processed in a few minutes). Once the discourse features are computed, the time spent by segmentation and reweighting algorithms remains static, introducing minimal overhead compared to the baselines. 

\paragraph{Discourse-driven Analysis on Factual Errors} In our analysis section, discourse analyses were carried out using the annotated portion of the released dataset, which is limited by the annotation quality and the dataset sizes. Yet, this is by far the only dataset that provides the sentence-level annotations on long document summarizations (i.e., \citet{krishna-etal-2023-longeval} released the fine-grained scores, but did not clarify how the spans annotations are collected in their document). We verify the effectiveness of portions of our linguistic-inspired method on other benchmarks, including \textsc{LongSciVerify} and \textsc{LongEval}. Future work would be to analyze and examine the discourse patterns in other domains, such as story summarization or further book-length summarization tasks \cite{chang2024booookscore, kim2024fables}. 

\paragraph{Generalize across Text Domains}
We tried to cover most of the recent publicly available factuality evaluation datasets for long document summarization, including \textsc{DiverSumm}, \textsc{LongSciVerify}, and \textsc{LongEval}. While most existing datasets consist of annotations collected from scientific article summaries, we introduce a novel annotated dataset, \textsc{LegalSumm}, in the legal domain to evaluate the robustness of our proposed approaches. This dataset is curated with careful annotation procedures to ensure quality (see Appendix \ref{appendix:legalsumm_detail}). Our experimental results, as shown in the last column of Table  \ref{tab:aggrefact_diversum_res}, demonstrate that our proposed approaches not only enhance the performances of baseline models but also surpass those strong LLM-based models by a large margin. 

\paragraph{Dependence on Discourse Parser Performance} Our experiments' validity and subsequent findings rely on the parsed discourse trees generated by a Rhetorical Structure Theory (RST) parser \cite{liu-etal-2021-dmrst}, following prior work \cite{adams-etal-2023-generating,pu-etal-2023-incorporating, kim2024threads}. It is important to note that parsed results may be sub-optimal given the challenges of complex hierarchical structures of long documents and the differences between the model's training corpora and our tested domains. We acknowledge that RST parsers are gradually evolving and posit that better RST parsing results can further boost the model's performance. However, major obstacles to their broader adoption are the lack of publicly available models and user-friendly user guidance. Researchers recently incorporated LLMs in discourse parsing and obtained better benchmarking performance in RST \cite{maekawa-etal-2024-obtain}. Unfortunately, no available inference code exists to parse documents beyond pre-compiled benchmark datasets. We look forward to utilizing more robust parsers in future work.

On long source documents, we notice that the parser failed on the MNW split of the DiverSumm, given their input is a concatenation of multiple individual news articles. We opt for first splitting the original document into articles and then successfully parsing them individually. Regarding paragraph-level discourse parsing, we are concerned that it may disrupt the discourse continuity at the document level (i.e., where the beginning of one paragraph is connected to the previous paragraph). Therefore, we leave this exploration for future studies. However, this approach might be viable and beneficial for summarizing extremely long documents, such as books, where the explicit division into chapters and sections could enhance the process.

\paragraph{Applications of Document Structures to Other Tasks}

Document structures can and have been utilized in different tasks, including coherence analysis \cite{liu-etal-2024-unlocking}, machine translation evaluation \cite{joty-etal-2017-discourse}, sentiment analysis \cite{KRAUS201965}, machine-generated text detection \cite{kim2024threads}, etc. While applying document structure and discourse analysis to hallucination detection is still an emerging area of research, we are keen to explore it further. We are also interested in extending this approach to other input sources, such as dialogue, by investigating the corresponding discourse structures unique to conversational data.

\section*{Ethical Statement}
Throughout the paper, we have referenced datasets and models used in our analyses and experiments, ensuring that they are openly available and do not pose concerns with the public release or usage of this paper. We acknowledge the use of Grammarly and ChatGPT-4o for correcting sentences that are less fluent but not for generating or drafting new content.

\bibliography{custom}

\appendix

\newpage

\newpage
\clearpage
\section{Discourse Analyses}
\subsection{Short Summary Analysis}\label{appendix:short_analysis}
\begin{table}[h!]
\scriptsize
\setlength\columnsep{1pt}
\begin{NiceTabular}{l|ccc}

\toprule
\textbf{Dataset} & \textbf{Size} &  \textbf{Gran} & \textbf{Error Tag} \\
\midrule

\textsc{AGU}\_\textsc{Cliff} &  300 & word & intrin./extrin./other/wld. knowl.\\
\textsc{AGU}\_Goyal'22 & 150   & span & intrins./extrin./other \\
\bottomrule
\end{NiceTabular}
\caption{Statistics of Sent/Span-level factual inconsistency datasets \textsc{AggreFact-Unified} (AGU) \cite{tang-etal-2023-understanding}. We report the size of doc-summary pairs (Size), the granularity of annotation (Gran), and the error labels (Error Tag).  }\label{tab:token_label_short}
\end{table}
We also conduct a discourse analysis on \textsc{AggreFac-United} \cite{tang-etal-2023-understanding}, as shown in Table \ref{tab:token_label_short}. This dataset includes BART and Pegasus summaries from \textsc{Cliff} \cite{cao-wang-2021-cliff} and Goyal'21 \cite{goyal-durrett-2021-annotating}.\footnote{\textsc{AggreFact-Unified} (\textsc{AGU\_Cliff}) includes additional error types such as \textit{comments}, \textit{other errors: noise, grammar} and \textit{world knowledge} (wld. knowl.)}  In the Goyal22 split of AGGREFACT-UNITED,
a total of 61 errors were detected. Intrinsic errors
are found to appear more often in satellite EDUs (18/31) with the attribution relation. Regarding extrinsic errors, the nucleus EDUs take the majority. We further analyzed the CLIFF dataset \cite{cao-wang-2021-cliff}, where span-level annotations of faithful errors are available. Out of 600 sentences, the parser failed to parse 131 summaries, likely due to their short lengths and simplistic structures. Therefore, our analysis focused on the 469 summaries that were successfully parsed. We observed that Elementary Discourse Units (EDUs) containing errors are more likely to appear at the bottom of the discourse tree.  These findings are similar to the long summary analysis in \S \ref{sec:discourse_analysis}. 


\subsection{Discourse Features}
\label{appendix:disocurse_feature_details}

Following prior work \cite{louis-etal-2010-discourse}, we analyze the nucleus-satellite penalty score (Ono penalty) \cite{ono-etal-1994-abstract}, the maximum depth (Depth score) \cite{Marcu1998tobuild}, and the promotion-based score \cite{Marcu1998tobuild} for sentence level. The penalty/score for
a sentence is computed as the maximum of the
penalties/scores of its constituent EDUs.
For the normalized version, instead of following \citet{louis-etal-2010-discourse}, who normalized them by the number of words in the
document, we opt to divide the scores by the maximum depth of the discourse tree, which similarly alleviates the scores' dependencies on document length. Below, we provide one example demonstrating the computation of each score (borrowed from \citet{louis-etal-2010-discourse}) and will release our code for reproduction purposes.

\begin{figure}[t!]
    \centering
    \includegraphics[width=0.9\linewidth]{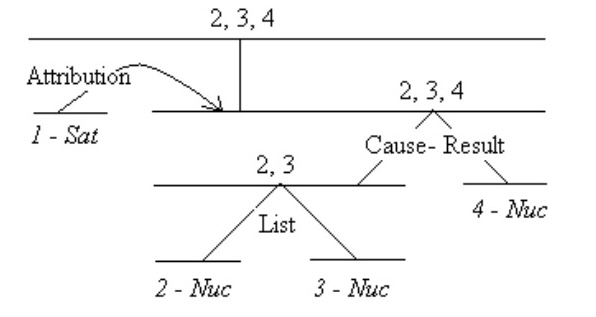}
    \caption{RST for the example sentence, and the salient units (promotion
set) of each text span are shown above the horizontal line, which represents the span.The example is taken from \citet{louis-etal-2010-discourse}.}
    \label{fig:discourse_example}
\end{figure}

\subsubsection{Example}
Here, we re-utilize the example from \citet{louis-etal-2010-discourse}, which is part of the RSTDT \cite{carlson2002rst} in Figure \ref{fig:discourse_example}, which contains four EDUs.

\textit{1. [Mr. Watkins said] 2. [volume on Interprovincial’s system is down about 2\% since January] 3. [and is expected to
fall further,] 4. [making expansion unnecessary until perhaps
the mid-1990s.]} 
\paragraph{Nucleaus-Satellite Penalty (Ono Penalty)}\cite{ono-etal-1994-abstract}: The spans of individual EDUs are represented
at the leaves of the tree. At the root of the tree, the
span covers the entire text. The path from EDU 1
to the root contains one satellite node. It is, therefore, assigned a penalty of 1. Paths to the root from
all other EDUs involve only nucleus nodes; subsequently, these EDUs do not incur any penalty. Thus, the Ono Penalty scores for EDU 1 to 4 are [1, 0, 0, 0].

\paragraph{Maximum Depth Score}
Below we cite the original texts from \cite{louis-etal-2010-discourse}.
\begin{quote}

\citet{Marcu1998tobuild} proposed the method to utilize the nucleus-satellite distinction, rewarding nucleus status instead of penalizing the satellite. He introduced the notion of \textit{promotion set}, consisting of
salient/important units of a text span. The nucleus is denoted as the more salient unit in the full span of
a mono-nuclear relation (i.e., in Elaboration, the satellite unit is to elaborate on the key information of the nucleus. Thus, the latter is more salient). In a multinuclear relation,
all the nuclei are salient units of the larger span.

For example, in Figure \ref{fig:discourse_example}, EDUs 2 and 3 participate in a multinuclear (List) relation. As a result,
both EDUs 2 and 3 appear in the promotion set of
their combined span (2-3). The salient units (promotion
set) of each text span are shown above the horizontal line which represents the span. At the leaves,
salient units are the EDUs themselves.

For the purpose of identifying important content, units in the promotion sets of nodes close to
the root are hypothesized to be more important
than those at lower levels. The highest promotion of an EDU occurs at the node closest to the
root, which contains that EDU in its promotion set.
The depth of the tree from the highest promotion
is assigned as the score for that EDU. Hence, the
closer to the root an EDU is promoted, the better
its score. Since EDUs 2, 3 and 4 are promoted all
the way up to the root of the tree, the score assigned to them is equal to 4, the total depth of the
tree. EDU 1 receives a depth score of 3.
\end{quote}

Thus, the final maximum depth score based on the promotion set for EDUs 1-4 are [3, 4, 4, 4]. 

\paragraph{Promotion Score}
In the same example, while EDUs 2, 3, and 4 all have a depth score of 4, EDUs 2 and 3 are promoted to the root from a greater depth than EDU 4. To account for the difference, \citet{Marcu1998tobuild} further introduced the promotion score, which is a measure of the number
of levels over which an EDU is promoted. For instance, EDU 2 is promoted by three levels, while EDU 4 is promoted by two levels. Thus,
EDUs 2 and 3 receive a promotion score of 3, while the score of EDU 4 is only 2. EDU 1, given that it is never promoted received scores of 0.

\paragraph{Discourse Tree Computation}
In Section \ref{sec:discourse_analysis} Table \ref{tab: subtree_struct}, we compute the tree depth as follows.
We use a string-matching system to construct a dictionary that aligns annotated sentences with EDU segments. For instance, in Figure \ref{fig:discourse_example}, the sentence is mapped to EDUs 1-4. We then compute the maximum depth of the discourse tree from the root node to the lowest leaf node, which would be 3 in this case. However, there may be cases where sentences are segmented into EDUs that are not gathered into a single node in the parsed discourse tree. In such instances, we employ the methods described in Section \ref{sec:discourse_analysis} to approximate the depth.

\section{LegalSumm Dataset}\label{appendix:legalsumm_detail}
We utilized a subset of the \textbf{CanLII Dataset} \cite{xu2021}, which consisted of 1,049 legal opinion documents with expert-written summaries.\footnote{Data obtained through an agreement with CanLII
\url{https://www.canlii.org/en/}}. We followed the setting from \citet{elaraby-etal-2024-adding}, where we consider the output of three different abstractive models in our annotation process: (1) \textbf{Finetuned LED-base} \cite{elaraby-litman-2022-arglegalsumm} which finetuned
the pre-trained longformer-encoder-decoder \cite{beltagy2020longformer} (LED) on the CanLII cases without additional information about the argument structure
of the document (2) \textbf{arg-LED-base}, which utilizes the LED model but includes the information about the argument units (Issues, Reasons, and Conclusions) in its training phase, and (3) \textbf{arg-aug-LED-base}, a model introduced in \citet{elaraby-etal-2023-towards} that can select a summary from multiple augmented versions of generated summaries based on its overlap with the input case’s predicted argument roles. 

\paragraph{Annotation Details}
We conducted evaluations with two voluntary legal experts from the research group, all of whom hold a J.D. degree and
possess at least four years of experience
in providing professional legal services. For each summary, the annotators are asked to select from four choices justifying the factual consistency of the model-generated summary with the reference summary and source article. They are also encouraged to provide free-text rationales justifying their selections.

\begin{figure*}
    \centering
    \includegraphics[width=0.9\linewidth]{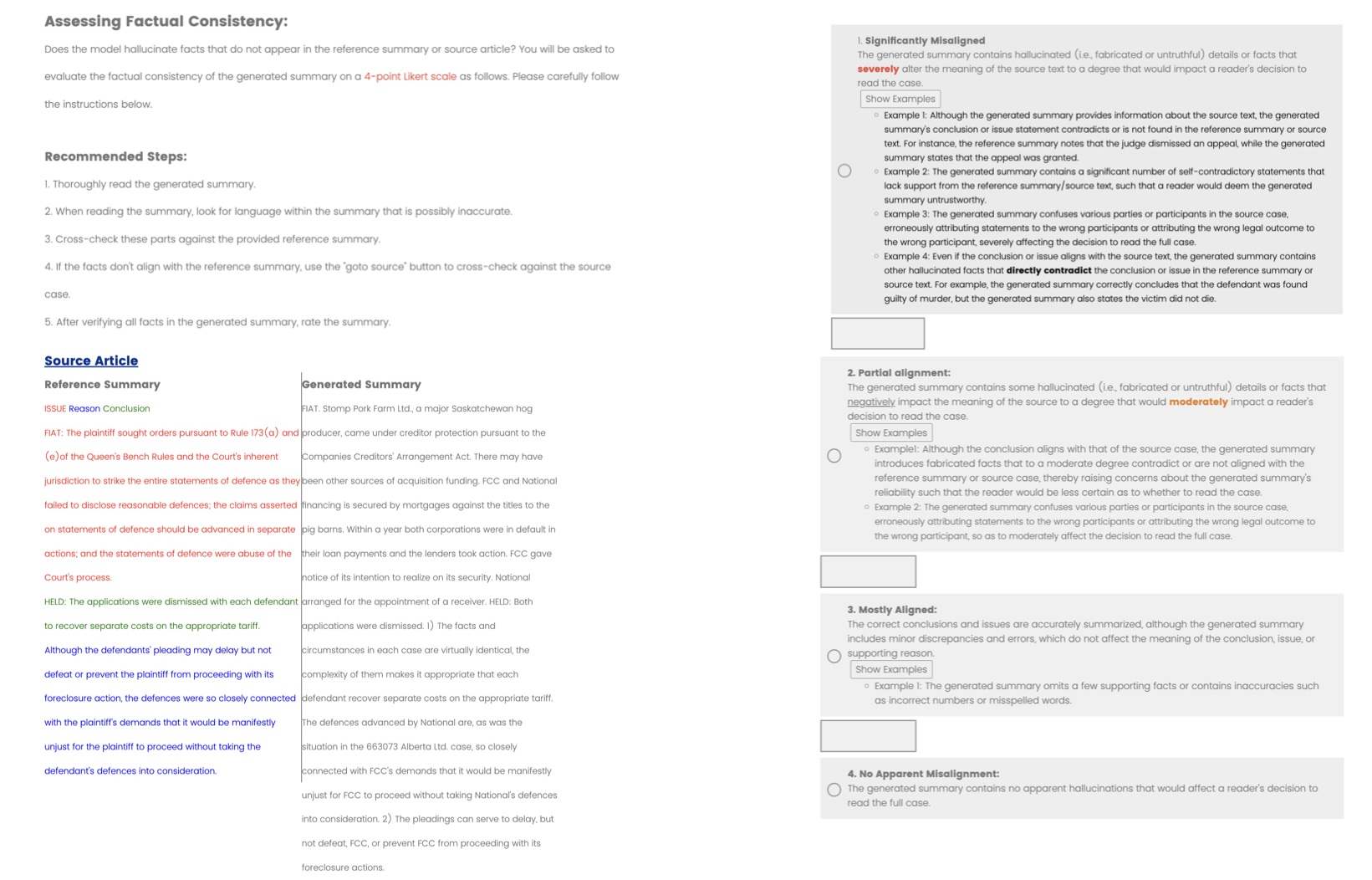}
    \caption{The annotation interface for LegalSumm. The left panel displays the instructions and the content to be annotated. Annotators are then prompted to select one of four options, as shown in the right panel.}
    \label{fig:annotation_guideline}
\end{figure*}

To guarantee the quality of annotation, we conducted multiple
sessions with annotators to refine the guidelines and continuously monitor the agreements. Ultimately, the first author and the two annotators held in-person sessions to resolve label inconsistencies. The labels remained unresolved in two cases as the annotators identified differing yet reasonable interpretations of the instructions.  We thus retain the average scores as is. To distinguish summaries with severe or moderate factual inconsistencies from those without, we computed the average of the two annotators' ratings and rounded based on a threshold of 2.
The annotation guideline is included in Figure \ref{fig:annotation_guideline}.

\section{Discourse Analysis on Fine-grained Error Types}\label{sec:appendix_discourse_test}
\begin{table*}[h!]
    \centering
    \scriptsize
    \begin{NiceTabular}{l|c|c|c|c|c|c|c|c}
    \toprule
    \textbf{RST features}     & \textbf{GramE} & \textbf{LinkE} & \textbf{OutE} & \textbf{EntE} & \textbf{PredE} & \textbf{CorefE} & \textbf{CircE} & \textbf{ALL Errors}\\
    Count & (83) & (35) & (48) & (117) & (15) & (9) & (13) & (320)\\
    
    \midrule
      Ono penalty   &  -1.166\textsuperscript{}& 1.855\textsuperscript{}& 0.621\textsuperscript{}& 1.647\textsuperscript{}& 0.730\textsuperscript{}& 0.215\textsuperscript{}& 1.627\textsuperscript{}& 1.606 (0.1089)\\
      Depth score& -5.218\textsuperscript{**}& -7.381\textsuperscript{**}& -4.628\textsuperscript{**}& -3.252\textsuperscript{**}& -2.002\textsuperscript{}& 0.214\textsuperscript{}& -0.565\textsuperscript{}& -8.249 (0.0000)\\
    Promotion score  & -6.519\textsuperscript{**}& -0.971\textsuperscript{}& -0.440\textsuperscript{}& 1.734\textsuperscript{}& -0.195\textsuperscript{}& 2.613\textsuperscript{*}& 0.629\textsuperscript{}& -0.828 (0.4083)\\
    \midrule
    Normalized penalty  &  -1.742\textsuperscript{}& 3.051\textsuperscript{**}& 0.695\textsuperscript{}& 1.990\textsuperscript{*}& 0.673\textsuperscript{}& -0.002\textsuperscript{}& 0.493\textsuperscript{}& 2.160 (0.0314)\\
      Normalized depth score & -6.689\textsuperscript{**}& -6.043\textsuperscript{**}& -4.823\textsuperscript{**}& -3.307\textsuperscript{**}& -1.731\textsuperscript{}& -0.153\textsuperscript{}& -1.986\textsuperscript{}& -9.084 (0.0000)\\
        Normalized promotion score &  -5.754\textsuperscript{**}& 0.487\textsuperscript{}& -0.322\textsuperscript{}& 1.796\textsuperscript{}& -0.087\textsuperscript{}& 2.206\textsuperscript{}& -0.218\textsuperscript{}& -0.303 (0.7617)  \\
    \bottomrule
      
    \end{NiceTabular}
    \caption{Two-sided t-test statistic of significant RST-based
features comparing unfaithful sentences to faithful
ones in \textsc{DiverSumm} annotated split. We report the
test statistics and significance levels. For fine-grained errors, we report the significant level in * (0.01 <= p-value <=0.05) and ** (p-value <=0.01). For All errors, we report the p-value in parenthesis.}
    \label{tab:rst_result_all_errors}
\end{table*}

\paragraph{Error Types} Relation
Error (PreE) is when the predicate in a summary sentence is inconsistent with respect to the document. Entity Error (EntE) is when the primary
arguments of the predicate are incorrect. Circumstance Error (CircE) is when the predicate’s circumstantial information (i.e., name or time) is wrong.
Co-reference error (CorefE) is when there is a pronoun or reference with an incorrect or non-existing
antecedent. Discourse Link Error (LinkE) is when
multiple sentences are incorrectly linked. Out of
Article Error (OutE) is when the piece of summary
contains information not present in the document.
Grammatical Error (GramE) indicates the existence
of unreadable sentences due to grammatical errors.


\paragraph{Fine-grained Error Analysis} In Table \ref{tab:rst_result_all_errors}, we demonstrate the breakdowns of fine-grained error types and report the t-test results on different discourse features.

\section{Example of Segmentation Failures}\label{appendix:segmentation_examples}
This section includes one example of the \textsc{AlignScore}'s chunking method that failed to preserve the document structure, while our discourse-inspired chunk addresses it.

\begin{figure*}[]
\centering
    \begin{subfigure}[b]{1\textwidth}
        \centering
        \includegraphics[width=1\textwidth]{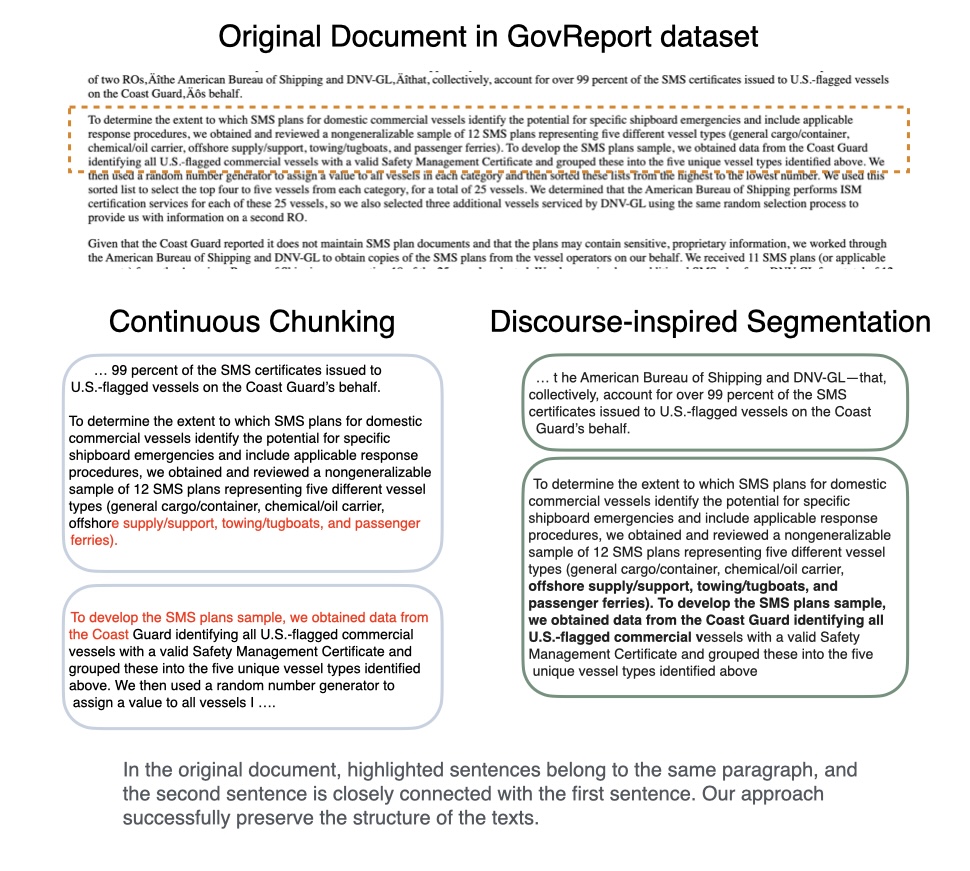}
        \caption{Example from GovReport of \textsc{DiverSumm}.}\label{fig:sub1}
    \end{subfigure}%
    \caption{Example of segmentation failures, left is the output of chunking method used in \textsc{AlignScore} and \textsc{MiniCheck}, right is the segments produced by our segmentation method.}\label{fig:segmentation_fault1}
\end{figure*}

\begin{figure*}[]
    
    \begin{subfigure}[b]{1\textwidth}
        \centering
        \includegraphics[width=1\textwidth]{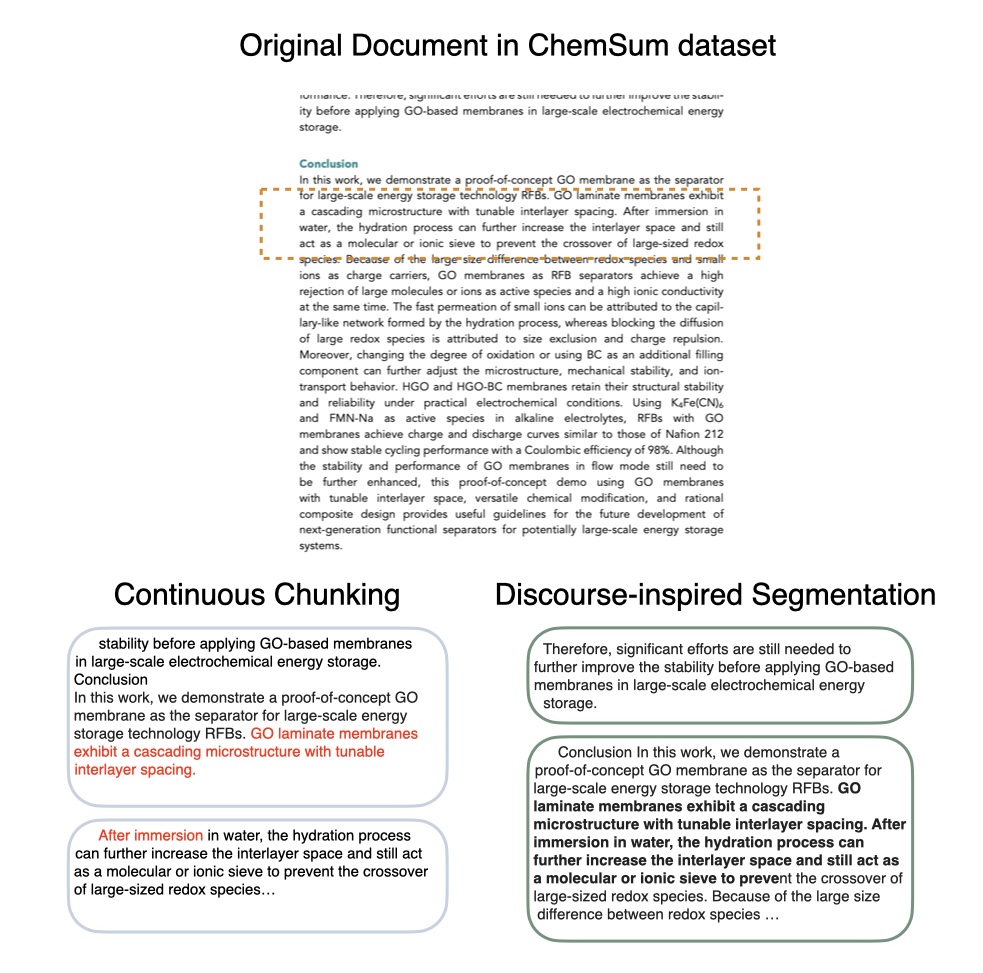}
        \caption{Example from ArXiv of \textsc{DiverSumm}.}\label{fig:sub2}
    \end{subfigure}
    \caption{Example of segmentation failures, left is the output of chunking method used in \textsc{AlignScore} and \textsc{MiniCheck}, right is the segments produced by our segmentation method.}\label{fig:segmentation_fault2}
\end{figure*}

For example, as shown in Figure \ref{fig:sub1}, the original document contains two consecutive sentences: "To determine the extent ..." and "To develop the SMS" (highlighted in the orange box). These sentences are meant to be read together and should not be separated.  However, the default chunking approach in \textsc{AlignScore} and \textsc{MiniCheck} breaks this continuity by placing them in two separate chunks, given the former chunk is large enough. On the contrary, our approach maintains the structural integrity of the documents, keeping the sentences connected as intended. Similarly, in Figure \ref{fig:sub2}, the conclusion section is separated into two chunks by the default chunking approach, while our method maintains them in a single chunk.

\newpage
\clearpage
\section{Implementation Details}\label{appendix:implementation_details}

\subsection{GPT4o Prompts}
We include our prompt for zero-shot factual consistency evaluation in Table \ref{tab:gpt4o_prompt}.
\begin{table*}
\scriptsize
    \begin{center}
        \begin{tabular}{l}
        \toprule
             Determine whether the provided claims are consistent with the corresponding document. Consistency in this context\\
             implies that
    all information presented in the claim is substantiated by the document.  If not, it should be \\
    considered inconsistent.\\
    \\
    Document: [DOCUMENT]\\
    Claims: [CLAIMS] \\
    Please assess the claim’s consistency with the document by responding with either "yes" or "no".\\
    The CLAIMs are ordered in the format of a dictionary, with \{ index: CLAIM \}. You will need to return the result in JSON format.\\
    For instance, for a CLAIMs list of 4 items, you should return \{0:yes/no, 1:yes/no, ...., 3:yes/no\}.\\
\\
    ANSWER: \\
            \bottomrule
        \end{tabular}
        \caption{Zero-shot factual consistency evaluation prompt for GPT4o.}\label{tab:gpt4o_prompt}
    \end{center}
\end{table*}

\subsection{Baselines}
\paragraph{AlignScore} (model size 355M) \cite{zha-etal-2023-alignscore} is an entailment-based model that has been trained on data from a wide range of tasks such as NLI, QA, and fact verification tasks. It divides the source document into a set of sequential chunks at sentence boundaries. For a multi-sentence summary, it predicts the max scoring value of all combinations of source chunk and target sentence, then returns the unweighted average of all sentences as the summary prediction. We follow the original setting by setting chunk size at 350 tokens and use the default model alingsocre\_large ckpt. The model outputs a score between 0 and 1. We conduct experiments on top of their released codebase \url{https://github.com/yuh-zha/AlignScore}.
\paragraph{MiniCheck-FT5} (model size 770M) \cite{tang2024minicheck} is an entailment-based fact checker built on flan-t5-large. It has been further fine-tuned on 21K datapoints from the ANLI dataset \cite{nie2019adversarial} and 35k synthesized data points generated in \cite{tang2024minicheck} on the tasks to predict whether a given claim is supported by a document. We follow the authors's setting and set the chunk size to 500 tokens using white space splitting. The output score is between 0 and 1. We use the released code repo from \url{https://github.com/Liyan06/MiniCheck}.

\paragraph{LongDocFactScore} \cite{bishop-etal-2024-longdocfactscore-evaluating} is a reference-free framework for assessing factual consistency. It splits source documents and the generated summary into sentences, then computes the pair-wise similarities by computing the cosine similarities of sentences (they use the sentence-transformers library initialized with the bert-base-nmli-mean-tokens model). Afterward, for each individual summary sentence,  K most similar source sentences are picked.  The method extracts the neighboring source document sentences of the selected sentences as context, then applies a metric  BARTScore to evaluate the score between source context and summary sentences. The overall summary score is an unweighted average of all sentences. We follow the authors' parameters setting and utilize their released code repo from \url{https://github.com/jbshp/LongDocFACTScore}.

\paragraph{InfUSE} (model size 60M) \citet{zhang-etal-2024-fine} uses a variable premise size and breaks the summary into sentences or shorter hypotheses. Instead of fixing the source context, it retrieves the best possible context to assess the faithfulness of an individual summary sentence by applying an NLI model to successive expansions of the document sentences. Similar to prior approaches, it outputs an entailment score for each summary sentence, and the summary-level score is the unweighted average. We follow their settings on \textsc{InfUsE} with summary sentences instead of \textsc{InfUsE\textsubscript{SUB}} as the authors only released the code for the former model. \textsc{InfUsE} outputs scores in the range 0-1. We use the author's released codebase from \url{https://github.com/HJZnlp/Infuse}.
\paragraph{GPT4o}
 We used the version of gpt-4o-2024-05-13; we set max\_tokens 100, sampling temperature at 0.7, and  top\_p as 1.0. We call the OpenAI API from \url{https://openai.com/api}. Given the lengthy summary, we prompted the LLM to assign a binary label (yes/no) to assess individual summary sentences' consistency with the original article. Then, we reported the percentile of ``yes'' answers as the summary-level rating. 

 \paragraph{BeSpoke-MC-7B}
 We harnessed the SOTA Llama-3.1-Bespoke-MiniCheck-7B (BeSpoke-MC-7B) released by Bespoke Labs. The model is fine-tuned from ``internlm/internlm2\_5-7b-chat'' \cite{cai2024internlm2} on the combination of 35K data points following the approach in MiniCheck \cite{tang2024minicheck}. We use the suggested code repo from \url{https://huggingface.co/bespokelabs/Bespoke-MiniCheck-7B}. To calculate the AUC score, we employed the raw probabilities returned by the code to determine sentence-level ratings, and we calculated the summary-level score as the unweighted average across all sentences.

\subsection{Machine Configuration for Models}\label{appendix:machine_config}
We use up to 4 NVIDIA RTX 5000 GPUs, each equipped with 16 GB VRAM,  for model inferences on our hardware. According to Lambda\footnote{\url{https://lambdalabs.com/service/gpu-cloud}} (RTX5000 is depreciated), a single NVIDIA Quadro RTX 6000 (the closest to our setting) GPU costs \$0.5 per hour and has 24 GB VRAM. Additionally, we loaded the Bespoke-MC-7B model on a single NVIDIA L40S GPU with 48 GB of VRAM, provided by the Pitt CRC computing cluster.

\section{Experimental Results}
\subsection{Discussion on Performance Compared to Strong Baselines}\label{appendix:more_discussion}
Our primary analysis focuses on discussing how the proposed approach can improve different baselines (we utilized three backbone baselines: rows 5, 11, and 17 with their improved versions) in Table \ref{tab:aggrefact_diversum_res}. We observe several baselines obtained the best performance on certain tasks and provide a more careful justification below: 

\paragraph{}
While the improvements may appear marginal in some baseline models, they are statistically significant and consistent across multiple datasets. The capabilities of baseline models and the characteristics of testbeds can also affect performance. For instance, as noted in Section \ref{sec:result}, dialogue-based inputs in QMS limit the effectiveness of discourse parsing (RQ2), while short summaries like XSUM minimize the impact of reweighting (RQ1). On longer datasets like AXV and CSM, gains are more substantial, with improvements of up to 7 points (row 14 vs. row 11 in AXV). This is comparable to, or even more significant than, prior work \cite{zhang-etal-2024-fine}, and it is common to observe varying levels of performance gains across different tasks \cite{tang-etal-2023-understanding,tang2024minicheck}.

\paragraph{LongDocFactScore} (LDFS) introduced the LongSciVerify (LSV) dataset (PUB and AXV), using a different annotation method by subsampling three sentences with human annotations for factuality. We conjecture this may lead to less accurate summary-level labels, favoring their metric, which utilizes the top-k sentence-level scores. Meanwhile, LDFS underperformed compared to most other baselines on AggreFact, QMS, AXV (from \textsc{DiverSumm}), and LongEval-PUB. In contrast, our approach outperformed LongDocFactScore on most other benchmarks (e.g., 86.32 vs. 65.36 on AXV), suggesting our approach is more robust and capable of handling different long document summarization datasets. While each baseline may excel in specific tasks, a more robust benchmarking dataset could better ensure fair comparisons for future research.

\paragraph{GPT-4o} GPT4o is utilized as a comparison between the SOTA LLMs (GPT4o models have unknown sizes but could be greater than known open-sourced LLMs with up to 405B) and our lightweight model (770M), which in the usual case, the LLMs can outperform baselines by noticeable margins \cite{tang2024minicheck}). In Table \ref{tab:aggrefact_diversum_res}, regarding the long document summarization datasets (from GOV in DiverSumm to LegalSumm), our models (rows 12, 13) outperformed GPT4o in 5 out of 6 test sets (the only exception is LongEval PUB). This confirmed that the discourse-inspired approaches are beneficial.

\paragraph{BeSpoke-MC-7B} is claimed to be the best fact-checking model publicly available on the LLM-AggreFact benchmark, which outperformed many other LLMs with bigger sizes. Compared to our proposed models, it performed better on QMS, XSM\textsubscript{AG}, LSV-AXV, and had the best performance on LongEval-PUB (similar pattern to GPT-4o). However, on other benchmarks, our discourse approaches still demonstrate their benefits (i.e., on \textsc{LegalSumm}, AlignScore + reweighting obtained 76.57 while BeSpoke-MC-7B only scored 55.81).

\subsection{Ablation Study}\label{appendix:ablation_study}
Table \ref{tab:larger_ablation} presents the ablation results of different discourse features on our baselines. We cover the long document summarization tasks starting from QMS in Table \ref{tab:aggrefact_diversum_res}.

\begin{table*}[ht!]
\small
\centering
\begin{NiceTabular}{lccccccc}
 \toprule
\multirow{1}{*}{\shortstack{\textbf{Model} }}
 &  \textbf{QMS} &  \textbf{GOV}  & \textbf{AXV}  & \textbf{CSM} & \textbf{LSV-PUB} & \textbf{LSV-AXV} & \textbf{LE-PUB}\\
\midrule

{MC-FT5 (SENT)} &  60.66 & 83.24 & 78.66 & 59.74 & 55.7 & 52.7 & 30.2 \\

{\hspace{3mm}\textit{+ subtree height}} & 60.21 & 84.55 & 79.09 & 60.55& 53.6 & 55.1 & 30.4\\
{\hspace{3mm}\textit{+ depth score}} & 60.51 & 83.65 & 78.90 & 59.90 & 55.7 & 53.8 & 33.3  \\
 re-weighting  & 60.36 & 84.75 & 79.38 & 60.06& 52.8 & 55.1 & 31.4 \\
\midrule 
\midrule
{AlignScore} & 56.48 & 87.02 & 77.46 & 61.03 & 54.9 & 53.9 & 36.9 \\

{\hspace{3mm}\textit{+ subtree height}} &52.91 & 87.29 & 81.15 &  60.47 & 51.7 & 55.4 & 34.1\\
{\hspace{3mm}\textit{+ depth score}} & 56.63 & 87.29 & 77.66 & 60.30 & 54.3 & 52.4 & 36.6  \\
 re-weighting  & 53.95 & 87.29 & 81.15 & 60.55 & 53.0 & 54.3 & 34.8 \\

\bottomrule
\end{NiceTabular}
\caption{Ablation results on long document datasets from \textsc{DiverSumm}, \textsc{LongSciVerify} and \textsc{LongEval}.}\label{tab:larger_ablation}
\end{table*}


\end{document}